\documentclass{vgtc}                          

\ifpdf%
  \pdfoutput=1\relax                   
  \usepackage{graphicx}                
  \DeclareGraphicsExtensions{.pdf,.png,.jpg,.jpeg} 
\else%
  \usepackage{graphicx}                
  \DeclareGraphicsExtensions{.eps}     
\fi%

\usepackage{microtype}                 
\PassOptionsToPackage{warn}{textcomp}  
\usepackage{textcomp}                  
\usepackage{mathptmx}                 
\usepackage{times}                    
        
\usepackage{cite}                      
\usepackage{tabu}                      
\usepackage{booktabs}                  

\onlineid{0}
\vgtccategory{Research}
\vgtcinsertpkg
\graphicspath{{figures/}{./}} 
\usepackage{array}
\usepackage{algorithmic}
\usepackage[ruled, vlined,linesnumbered,commentsnumbered]{algorithm2e}
\usepackage[caption=false,font=footnotesize]{subfig}

\usepackage{amssymb}
\usepackage{amsmath} 

\usepackage{amsfonts}
\usepackage{tabularx,ragged2e}

\newcommand {\mm}[1] {\ifmmode{#1}\else{\mbox{\(#1\)}}\fi}

\DeclareMathAlphabet{\mathcal}{OMS}{cmsy}{m}{n}

\newcommand{\Espace}        {\mm{\mathbb{E}}}

\newcommand{\Scal}        {\mm{{\mathcal S}}}
\newcommand{\Acal}        {\mm{{\mathcal A}}}
\newcommand{\Pcal}        {\mm{{\mathcal P}}}
\newcommand{\Rcal}        {\mm{{\mathcal R}}}
\newcommand{\Ncal}        {\mm{{\mathcal N}}}

\newcommand{\cmx}        {\mm{\chi^*}}

\newcommand{\RNC}        {\mm{\mathrm{R_{NC}}}}
\newcommand{\RNO}        {\mm{\mathrm{R_{NO}}}}
\newcommand{\RNE}        {\mm{\mathrm{R_{NE}}}}
\newcommand{\RNA}        {\mm{\mathrm{R_{NA}}}}

\newcommand{\etal}{\textit{et al.}}

\newcommand{\myarea}  {\mbox{area}}

\newcommand{\tool}{\textsf{MARLL}}

\newcommand{\para}[1]        {\vspace{2mm}\noindent{\textbf{#1}}}

\newcommand{\denselist}{\vspace{-5pt} \itemsep -3pt\parsep=-1pt\partopsep -3pt}

\usepackage{amsmath,hyperref}
\hypersetup{colorlinks=true}

\def\equationautorefname~#1\null{%
  Equation~(#1)\null
}

\usepackage{hyperref}

\title{Interpreting Graph Drawing with Multi-Agent Reinforcement Learning}

\author{Ilkin Safarli\thanks{e-mail: ilkin@sci.utah.edu} 
\and Youjia Zhou\thanks{e-mail: zhou325@sci.utah.edu}
\and Bei Wang\thanks{e-mail: beiwang@sci.utah.edu}}
\affiliation{\scriptsize Scientific Computing and Imaging Institute, University of Utah}

\abstract{
Applying machine learning techniques to graph drawing has become an emergent area of research in visualization.
In this paper, we interpret graph drawing as a multi-agent reinforcement learning (MARL)  problem. 
We first demonstrate that a large number of classic graph drawing algorithms, including force-directed layouts and stress majorization, can be interpreted within the framework of MARL. 
Using this interpretation, a node in the graph is assigned to an agent with a reward function. 
Via multi-agent reward maximization, we obtain an aesthetically pleasing graph layout that is comparable to the outputs of classic algorithms.  
The main strength of a MARL framework for graph drawing is that it not only unifies a number of classic drawing algorithms in a general formulation, but also supports the creation of novel graph drawing algorithms by introducing a diverse set of reward functions.

} 

\begin{document}

\maketitle

\section{Introduction}
Graphs are widely used to model complex relational data such as computer networks, brain connectomics, protein-protein interactions, and communication patterns within social media. 
To visualize graphs, various techniques have been developed in the graph drawing and information visualization literature (see e.g.,~\cite{HermanMelanconMarshall2000,TarawanehKellerEbert2012,GibsonFaithVickers2013} for surveys).

Node-link diagrams are perhaps one of the most popular and intuitive graph visualization methods, where nodes  are represented as points and edges as lines. 
However, drawing a graph with an arbitrary node-link diagram, on its own, does not necessarily lead to insight into its underlying data. 
The layout of nodes  and edges strongly influences how a graph and its underlying data are understood. 
Therefore, an important point in graph drawing is ``not simply drawing a graph but how the graph is drawn"~\cite{GibsonFaithVickers2013}.
Depending on the structure of the data and the nature of the application domain, various styles of layouts, such as force-directed, circular, orthogonal, and hierarchical, have been developed. 
Since no universal layout algorithm exist that can produces good results for all types of graphs, we ask a different question instead: \emph{Can we put a large number of classic graph drawing algorithms in a unifying framework}?
 
Even though many layout algorithms have been developed since Tutte~\cite{Tutte1960} first introduced his barycenter method in 1960, the graph drawing problem remains an active area of research. 
Recently, applying machine learning techniques to graph drawing has started to gain traction (e.g.~\cite{KwonCrnovrsaninMa2018, KwonMa2020, WangJinWang2020}). 
Many of these approaches rely on classic graph drawing algorithms to generate a training dataset, and then utilize machine learning models to learn from the training dataset. 
However, the training process can be time-consuming, its performance relies heavily on the similarity between the training and testing datasets, and the models typically do not generalize for different graphs and different layouts (e.g.~\cite{KwonMa2020}). 
We take a different perspective: \emph{What are the different ways classic graph drawing algorithms can be reinvented or transformed using machine learning?}  

In this paper, instead of a deep-learning-flavored approach (e.g.~\cite{KwonMa2020,WangJinWang2020}) that relies  on training datasets, we take a broader view of learning and focus on the paradigm of reinforcement learning, where we learn to ``choose the correct actions based on the outcomes of previous actions in similar situations"~\cite{Case2019}.
Specifically, we interpret the graph drawing problem as a multi-agent reinforcement learning (MARL) problem. 

We first demonstrate that a large number of classic graph drawing algorithms, including force-directed layouts, stress majorization, and incremental layouts, can be interpreted within the framework of MARL. 
We achieve this interpretation by assigning agents to nodes  with reward functions that are derived from classic layout algorithms.
An aesthetically pleasing layout that is comparable to the classic algorithm is obtained via multi-agent reward maximization.
We then propose a set of novel graph drawing algorithms by introducing a diverse set of reward functions. 
To the best of our knowledge, this is the first time reinforcement learning is utilized in graph drawing.

We investigate the effectiveness of the proposed framework through both qualitative and quantitative evaluations based on established metrics.
We develop a tool call {\tool} (pronounce as ``Mar Roll", which stands for Multi-Agent Reinforcement Learning  Layouts) to demonstrate the utility of our MARL approach. 
{\tool} is available via Github\footnote{https://github.com/kinimesi/marl-layout-demo}. 
It implements a set of classic layout algorithms and their MARL counterparts and allows the comparison of their resulting layouts across synthetic and real world datasets. 

In summary, our contributions include:
\begin{itemize}\denselist
    \item A novel, unifying framework that interprets a large number of classic graph drawing algorithms with MARL.
    \item A method for recombining existing and deriving new layout algorithms based on a diverse set of reward functions.
    \item A demonstration of the effectiveness of our method via qualitative and quantitative evaluations.
    \item An open-source tool that implements and demonstrates various classic and MARL layouts that facilitates visual comparisons.
\end{itemize}
\section{Related Work}
\label{sec:related-work}

A number of studies have applied machine learning to graph drawing, see~\cite{VieiraNascimentoSilva2015} for a survey.
We classify these studies based on the models they employ, namely, evolutionary learning, neural networks/deep learning, and other approaches. 
We also give a brief overview of related work in reinforcement learning. To the best of our knowledge, this is the first time reinforcement learning is used to study graph drawing algorithms. 

\para{Evolutionary learning.} 
Evolutionary algorithms are one of the early machine learning approaches for graph drawing. 
The positions of vertices in a graph layout represent the genotypes of individuals. 
The fitness function is a weighted sum of functions that measures how well a drawing accommodates certain aesthetic criteria~\cite{SuarezSebagRodriguez1999}.
Based on a fitness function, best candidate layouts are selected at each step. 
Crossover and mutation processes are then applied to the candidates to generate a new layout until the layout converges. 

The approaches that use evolutionary learning can be further classified into two categories: independent from or dependent on user interactions. 
Approaches in the first category automatically compute the layout of a given graph without any user interaction (e.g.,~\cite{Tettamanzi1998,KosakMarksShieber1991,SponemannDuderstadtHanxleden2014}). 
The fitness function encodes various aesthetic criteria, such as minimizing node overlaps (e.g.,~\cite{KosakMarksShieber1991,KosakMarksShieber1994}), minimizing the number of edge crossings (e.g.,~\cite{KosakMarksShieber1991,Tettamanzi1998}), and maximizing the uniformity of edge lengths (e.g.,~\cite{Tettamanzi1998,KosakMarksShieber1994}), see~\cite{SuarezSebagRodriguez1999} for a survey.

Many approaches in the second category follow a similar pipeline via evolutionary learning with user interactions (e.g.~\cite{Masui1994,SuarezSebagRodriguez1999,BarbosaBarreto2001,SponemannDuderstadtHanxleden2014}): users are first asked to rate a set of graph layouts; a fitness function that reflects their preferences is learned based on user feedback, and then a layout is produced from the fitness function. 
In the work by Masui~\cite{Masui1994}, users provided the system with pairs of good and bad layout examples, and the system inferred the fitness function from these examples, which then evolved via genetic programming. 
Rosete-Suarez \etal~\cite{SuarezSebagRodriguez1999} built a system that acquires user preferences either from user evaluations or from user actions.

\para{Neural networks for graph drawing.}
Cimikowski and Shope~\cite{CimikowskiShope1996} presented a parallel neural network algorithm for minimizing the number of edge crossing of nonplanar graphs in a linear embedding. 
Each edge in the graph is associated with an ``up" or "down" neuron, which represents an embedding of the edge in the upper or lower half-plane~\cite{CimikowskiShope1996}.
Two kinds of forces -- excitatory and inhibitory forces -- push the neurons iteratively to a state that minimizes the number of edge crossings~\cite{VieiraNascimentoSilva2015}. 
Their system does not build a neural network for learning purpose, but rather it models the graph structure as a neural network coupled with an energy system~\cite{VieiraNascimentoSilva2015}. 

Wang \etal~\cite{WangJinWang2020} proposed a graph-LSTM-based model to learn and predict the drawing of an input graph. 
LSTM (long short-term memory) is a popular version of recurrent neural networks (RNNs). 
Using a set of layout examples as the training dataset, a LSTM model was trained to learn the characteristics of the training dataset and its corresponding algorithm-specific parameters for determining  desirable graph drawings. 
In particular, direct connections were added between different LSTM cells to explicitly model the topological structure of input graphs. 
Given a new input graph, the trained model was then used to generate a graph drawing with visual properties similar to those of the training dataset. 

Kwon and Ma~\cite{KwonMa2020} introduced a deep generative model based on VAEs (variational autoencoders)~\cite{KingmaWelling2014} that systematically visualizes a graph in diverse layouts.
They present an encoder-decoder architecture to learn a VAE from a collection of example layouts, where the encoder represents training examples in a latent space and the decoder generates layouts from the latent space.
A what-you-see-is-what-you-get (WYSIWYG) interface was constructed by mapping the generated samples onto the 2D latent space.
Such an interface allowed users to navigate and generate among various representative layouts without adjusting parameters of layout methods~\cite{KwonMa2020}. 
However, although a trained model could generate different layouts for the same input graph, a new model needs to be trained for each new input. 
Generalizing such a model to an unseen graph remains challenging. 

\para{Other machine learning approaches for graph drawing.}
Kwon \etal~\cite{KwonCrnovrsaninMa2018} proposed a machine learning approach to the graph drawing problem that gives the user an idea of ``what a graph would look like'' in a particular layout.
Given an input graph, they did not compute its layout directly.
Instead, they found the most topologically similar graph to the input graph using graphlet frequency, and then displayed its pre-computed layout.
Their method can be quite useful, especially when an input graph is large and the user is interested only in an approximated layout. 
However, their method needs a training dataset that includes the layouts of many graphs and their quality metrics, which require a considerable amount of time to generate. 

\para{Reinforcement learning.}
A foundational idea underlying theories of learning and intelligence is \emph{learning from interaction}, and reinforcement learning (RL) is its corresponding computational approach~\cite{SuttonBarto2012}. 
There have been many recent advances in applying RL to well-known sequential decision-making problems; see~\cite{SuttonBarto2012} for an introduction and~\cite{KaelblingLittmanMoore1996,ZhangYangBasar2019} for surveys. 
Successful applications of RL include the AlphaGo system, which defeated a professional human Go player~\cite{SilverHuangMaddison2016, SilverSchrittwieserSimonyan2017}, robotics control~\cite{KoberPeters2014}, playing card games~\cite{BrownSandholm2017}, and autonomous driving~\cite{Shalev-ShwartzShammahShashua2016}.
RL systems are divided into two categories, single-agent RL and multi-agent RL.
As the name suggests, they differ by the number of agents operating in the system and the associated RL algorithms. 
We give some technical background on these systems in~\autoref{sec:background}.  
To the best of our knowledge, RL has not been applied in graph drawing as presented in this paper.

\section{Technical Background}
\label{sec:background}

\subsection{Graph Drawing}

An unweighted, undirected graph $G = (V, E)$ consists of a node  set $V$ and an edge set $E \subseteq V \times V$ connecting pairs of nodes. 
Let $|V| = n$ and $|E| = m$.  
The problem of graph drawing is to find a graph layout -- a set of coordinates $p_v$ for each node  $v \in V$ -- such that the drawing is aesthetically appealing~\cite{Hu2006}. 
In this paper, we consider 2D graph drawings where $p_v \in \mathbb{R}^2$. 
We also assume that edges are drawn as straight lines, and nodes have uniform sizes and circular shapes.

\para{Force-directed layout.}
Force-directed graph layout is a class of algorithms that is widely used for  simple undirected graphs~\cite{Kobourov2012,GibsonFaithVickers2013}. 
The graph is modeled as a physical system where nodes are attracted and repelled based on various force formulations.

Eades's spring-embedder inspired many force-directed layout algorithms~\cite{Eades1984}.
He proposed a mechanical model in which the nodes are represented with steel balls and the edges with springs.
The nodes are placed with a random configuration and released. 
The force on each node  in the system is the sum of repulsive forces from all nodes and attractive forces from adjacent nodes.
A layout is obtained once the system reaches equilibrium.

\para{FR layout.} Fruchterman and Reingold presented a modification of the spring-embedder model of Eades with increased speed and simplicity~\cite{FruchtermanReingold1991} (referred to as a FR layout in this paper). Their principles for graph drawing remain applicable even today: ``nodes connected by an edge should be drawn near each other" and ``nodes should not be drawn too close to each other"~\cite{FruchtermanReingold1991}. 
Fruchterman and Reingold were concerned with aesthetic criteria, including even node  distribution, minimized edge crossings, uniform edge lengths, inherent symmetry, and conforming the drawing to a given frame.
In their model, the nodes are represented as atomic particles or celestial bodies that ``exert attractive and repulsive forces on one another" to induce movement~\cite{FruchtermanReingold1991}.

First, the attractive forces and repulsive forces between a pair of nodes $u$ and $v$ are calculated with 
\begin{align}
f_a(u,v) & = \frac{d(u,v)^2}{k}, \label{eq:FR-a}\\  
f_r(u,v) & = \frac{-k^2}{d(u,v)}, \label{eq:FR-b}
\end{align}
where $d(u,v):=||p_u -  p_v||$ is the distance. 
\autoref{eq:FR-a} is computed for $(u,v) \in E$ and \autoref{eq:FR-b} is for all pairs of nodes. 
$k$ is the optimal distance between nodes;  it is calculated as $k = C\sqrt{\frac{\myarea}{|V|}}$, where $\myarea$ is the size of the drawing canvas and the constant $C$ is found experimentally. 
Then, the notion of temperature is added to control the displacement of nodes, so that as the layout becomes better, the adjustments of the nodes become smaller with respect to the decreasing in temperature~\cite{Kobourov2012}.

\para{DGC layout.}
Dogrusoz \etal~\cite{DogrusozGiralCetintas2009} proposed another spring-embedder variant for laying out general graphs with non-uniform node sizes and compound nodes (referred to as a DCG layout here).
Their force model also consists of both attractive and repulsive forces, calculated as 
\begin{align}
f_a(u,v) & = \frac{(\lambda - d(u,v))^2}{\zeta},\\
f_r(u,v) &= \frac{\mu}{d(u,v)^2}
\end{align}
where $\lambda$ is the ideal edge length, $\zeta$ is the elastic constant of the edge, and $\mu$ is the repulsion constant. 
Similar to Fruchterman and Reingold, they also limited the maximum node displacement based on a temperature cooling schema.

\para{Stress majorization.}
Gansner \etal~\cite{GansnerKorenNorth2005} proposed to perform energy optimization with stress majorization.
Their stress function is defined as 
\begin{align}
E = \sum_{u<v} w_{u,v} (d(u,v)- d'(u,v))^{2}
\end{align}
where $d'(u,v)$ is the graph-theoretic distance between $u$ and $v$, and $w_{u,v}$ is the normalization constant, which equals $d'(u,v)^{-2}$. 
This stress function can be globally optimized via majorization (a method from multidimensional scaling), which is guaranteed to converge~\cite{Kobourov2012}.
This approach can also improve the layout quality and running time in practice~\cite{GansnerKorenNorth2005}.
Zheng \etal~\cite{ZhengPawarGoodman2019} used stochastic gradient descent to minimize the same stress function.

\subsection{Multi-Agent Reinforcement Learning}
We review the technical formulations for finite single-agent reinforcement learning (SARL) and its multi-agent counterpart following the notations from the survey of Zhang \etal~\cite{ZhangYangBasar2019} with minor modifications based on the work of Busoniu \etal~\cite{BusoniuBabuskaDeSchutter2008}. 

\para{SARL.}
In a dynamic environment, an RL agent is an algorithmic component that learns by interacting with the environment via trials and errors~\cite{KaelblingLittmanMoore1996}. 
An agent senses the state of the environment and takes an action; its purpose is to reach its goal by taking actions that maximize its associated reward~\cite{SuttonBarto2012}. 
Such a process can be modeled as a Markov decision process (MDP). 

Formally, a SARL system can be formulated as an MDP via a tuple $(\Scal, \Acal, \Pcal, \Rcal, \gamma)$~\cite{ZhangYangBasar2019}:  
\begin{itemize} \denselist
\item $\Scal$ is the \emph{state space} that contains states of the environment. 
\item $\Acal$ is the \emph{action space} consisting of actions available to the agent. 
\item $\Pcal(s,a,s')$ is the \emph{transition probability} from any state $s \in \Scal$  to any state $s' \in \Scal$ for any given action $a \in \Acal$.
\item $\Rcal(s,a,s')$ is the \emph{reward function} that determines the immediate reward received by the agent after transitioning from $s$ to $s'$ with action $a$.
\item $\gamma$ is the \emph{discount factor} that balances between the instantaneous and future rewards.
\end{itemize}
The behavior of the agent is described by its policy $\pi$, which is a mapping from the state space $\Scal$ to the distribution over the action space $\Acal$ that describes how the agent chooses its actions given the state~\cite{BusoniuBabuskaDeSchutter2008}. 
$\Pcal$, $\Rcal$, and $\pi$ can be deterministic or stochastic. 
In a deterministic model, the next state,  the reward, and the policy are completely determined by the current state-action pair. 
In the stochastic model, they are drawn from distributions. 
We describe the stochastic model here. 

The goal is to maximize the expected \emph{discounted accumulated reward} by finding a policy $\pi$~\cite{ZhangYangBasar2019},  
\begin{align}
\Espace \bigg[\sum_{t\geq 0}\gamma^t \Rcal(s_t,a_t,s_{t+1}) 
{\,\bigg|\,} a_t\sim\pi(s_t),s_0
 \bigg], 
\end{align}
where $a_t \sim \pi(s_t)$ is the action drawn from the policy, and the expectation is taken over $s_{t+1} \sim \Pcal(s_t, a_t, s_{t+1})$. 

One way to achieve this maximization is by computing an optimal \emph{$Q$-function} (i.e.,~action-value function) for a state-action pair $(s,a)$ by capturing its expected return given the policy $\pi$~\cite{BusoniuBabuskaDeSchutter2008}:  
\begin{align}
Q_\pi(s,a)=\Espace\bigg[\sum_{t\geq 0}\gamma^t \Rcal(s_t,a_t,s_{t+1})
{\,\bigg|\,} a_t\sim\pi(s_t), a_0 = a, s_0 = s
\bigg].
\end{align} 
Various RL methods have been developed to find a good estimate of the optimal Q-function.
The (approximate) optimal policy can then be obtained by taking the $\epsilon$-greedy action of the Q-function estimate. 
One of the most popular methods for estimating the optimal Q-function is the Q-learning algorithm~\cite{WatkinsDayan1992}.
For a given agent that takes action $a$ from state $s$ to state $s'$ and gets a reward $r = \Rcal(s,a,s')$, the algorithm adjusts the estimated value of Q-function as 
\begin{equation}
\hat{Q}_\pi(s,a) \leftarrow (1-\alpha)\hat{Q}_{\pi}(s,a)+\alpha\big[r+\gamma\max_{a'}\hat{Q}_{\pi}(s',a')\big]
\label{eq:qFunction}
\end{equation}
where $\alpha$ is the learning rate, and $\gamma$ is the discount factor. 
The approximate optimal policy is obtained by taking the greedy action of the Q-function estimate~\cite{ZhangYangBasar2019}.
That is, an agent picks for every state $s$ the action $a$ with the highest value of the Q-function~\cite{BusoniuBabuskaDeSchutter2008}. 

\para{MARL.}
Using a generalized version of the MDP, a MARL system can be formulated as a Markov game (MG). 
It is specified by a tuple $(\Ncal,\Scal,\{\Acal^i\}_{i\in\Ncal},\Pcal,\{\Rcal^i\}_{i\in\Ncal},\gamma)$~\cite{ZhangYangBasar2019}. 
$\Ncal=[1,\cdots,N]$ denotes the set of $N>1$ agents. 
$\Acal^i$ and $\Rcal^i$ are the state space and the reward function of agent $i$, respectively. 

MARL systems are mainly divided into three main categories: fully cooperative, fully competitive, and mixed systems~\cite{BusoniuBabuskaDeSchutter2008, ZhangYangBasar2019}.
We implement the fully cooperative version in this paper, as the name suggests, all the agents share the same reward function and Q-function. It would be interesting to explore fully competitive and mixed systems in the future and observe their effect on graph layouts.

\section{Method}
\label{sec:method}

Our proposed approach models and thus interprets a number of graph drawing algorithms with MARL. 
To the best of knowledge, this is the first time RL is utilized in graph drawing. 

\subsection{Model Graph Drawings with MARL}
We interpret the graph drawing problem with MARL by assigning an agent to each node.
During each step of an iterative process, the agents can take actions and move toward the direction within the environment that maximizes their reward function. 
A graph layout is obtained via reward maximization. 

\begin{figure}
    \centering
    \includegraphics[width=0.98\columnwidth]{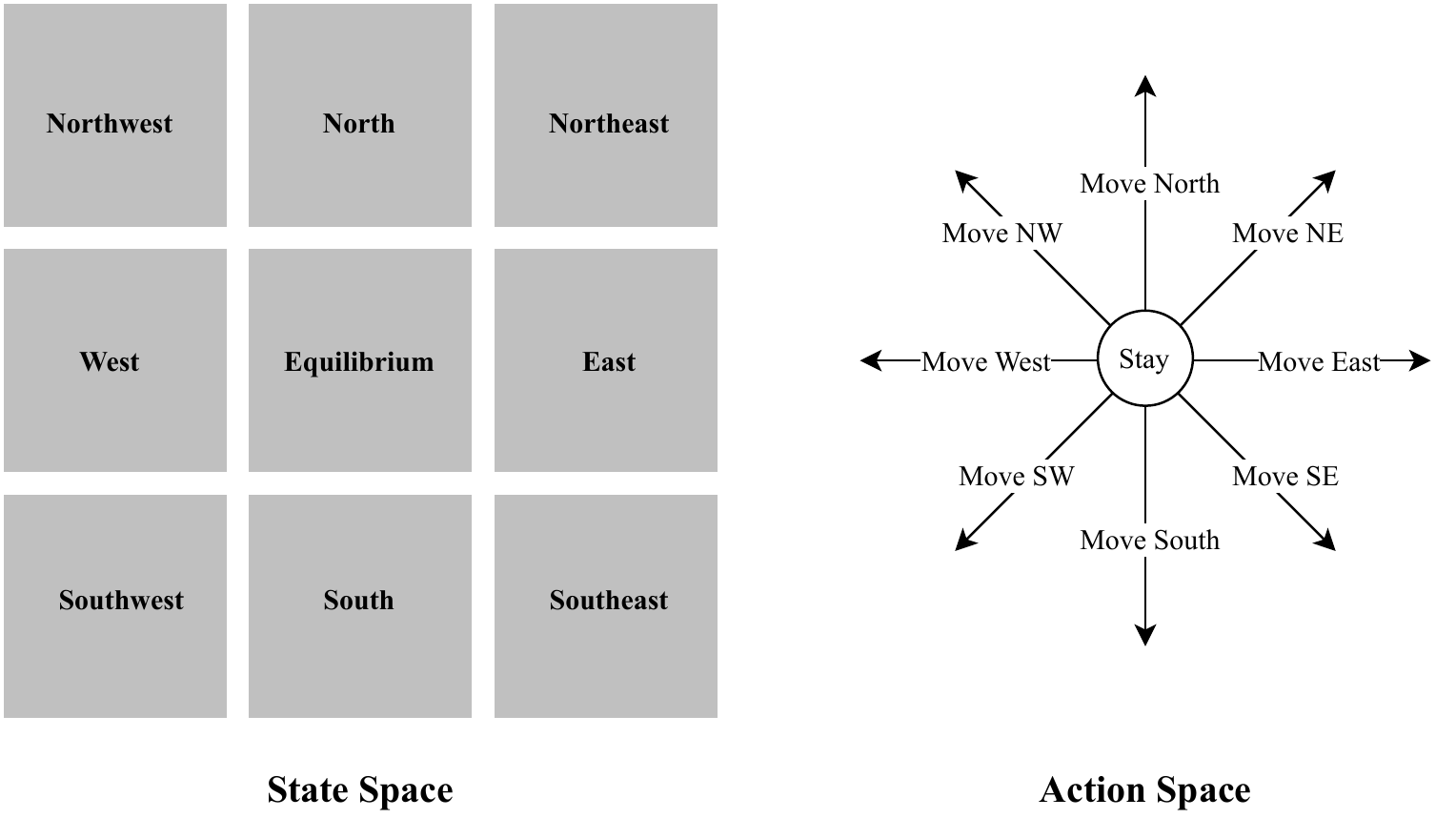} 
    \vspace{-2mm}
    \caption{The state and action space of a given agent. An agent can take any action from any state.}
    \label{fig:stateGrid}
\end{figure}

\para{States and actions.}
In our model, we divide the state space of a given agent into a $3 \times 3$ grid as shown in \autoref{fig:stateGrid}.
There are in total nine states.
An agent can move from any one of the nine total states to any other state by taking the relevant action. 
The action space consists of nine actions accordingly: stay (do not move), move north, south,  east, west, northeast, northwest,  southeast, and southwest, respectively. 
All actions are possible from any state.  

\para{Reward function.}
The reward function is the feedback mechanism for an agent that defines the consequences of the agent's actions and facilitates learning by interaction~\cite{SuttonBarto2012}.
In many force-directed and energy-based graph drawing algorithms, the  equilibrium state is obtained by minimizing the energy or the  stress of the system. 
In our model, we formulate the energy or the stress via a reward function. 
Suppose we assign an agent to a node in the graph. 
The difference of the force or energy (stress) on each node before and after the action of an agent is defined to be the reward function for a state-action pair. 
Specifically, for force-based formulation, the reward function for an agent $v$ at time (i.e.,~iteration) $t$ is defined as
\begin{equation}
\Rcal^{v}(t) = \|F^v{(t-1)}\| - \|F^v{(t)\|},
\label{eq:rewardFunction}
\end{equation}
where $F^v(t)$ is the total force on node $v$ at time $t$, $||\cdot||$ is the vector norm.
For energy-based formulation, the reward function is
\begin{equation}
\Rcal^{v}(t) = E^v{(t-1)} - E^v{(t)}.
\label{eq:rewardFunction2}
\end{equation}

The goal of any agent is to maximize its reward by moving the corresponding node around according to a policy that is learned by interacting with the environment.
The formulations in~\autoref{eq:rewardFunction} and~\autoref{eq:rewardFunction2} provide a general framework to interpret a number of graph drawing algorithms that are based on physical simulations. 
In~\autoref{sec:interpretation}, we discuss how different algorithms can be interpreted with this formulation.

\para{Finding optimal policy.}
In MARL, Q-learning is an iterative approximation procedure that estimates the optimal value function~\cite{WatkinsDayan1992,BusoniuBabuskaDeSchutter2008}.
For a given agent, the Q-function is given in~\autoref{eq:qFunction}. 
At each iteration, the action of an agent is selected based on an $\epsilon$-greedy policy~\cite{BusoniuBabuskaDeSchutter2008} (reviewed in~\autoref{sec:background}).  
If the associated reward of an action is positive, then the action is taken by moving the node to its new position.
If the reward is negative, then an action is accepted with a small probability to prevent the algorithm from becoming stuck at a local minimum (similar to a Metropolis–Hastings algorithm). 
The learning process continues until the system converges.

\para{Convergence criteria.}
We consider the convergence of a Q-learning algorithm as the convergence of its underlying graph layout. 
We demonstrate via experiments in~\autoref{sec:evaluation} that at convergence, we arrive at an aesthetically pleasing graph layout.  
Our model has four criteria for convergence.
The algorithm terminates whenever one of these criteria is reached. 
The first criterion $M$ is simply the total number of (allowable) iterations.  
 
Second, if the average node displacement is less than a predefined threshold, then the algorithm terminates.
The \emph{average node displacement} $A$ at time $t$ is given by 
$
A(t) = \frac{\sum_{v\in V}\lVert p_v{(t)} - p_v{(t-1)}\rVert}{n}, 
$
where $n$ is the number of nodes, and $p_v{(t)}$ is the position of node $v$ at time $t$. $\|\cdot\|$ denotes the Euclidean norm.  

Third, if the node displacement rate is smaller than a predefined threshold, then the algorithm terminates.
The \emph{displacement rate} $\Delta A$ at time $t$ is calculated by
$
\Delta A(t) = \|A(t) - A(t - 1)\|. 
$

Finally, we utilize the stress ratio as a termination criterion according to Gansner \etal~\cite{GansnerKorenNorth2005}, when the reward function is based on the energy of the system. 
That is, the algorithm terminates if the stress ratio falls below a predefined threshold.  
The \emph{stress ratio} $\delta E$ at time $t$ is given as
 $  \delta E(t) = \frac{E(t)- E(t-1)}{E(t)}, $
where $E(t)$ is the total energy of the system at time $t$. 

\subsection{Interpret Classical Layouts With MARL}
\label{sec:interpretation}

In \autoref{eq:rewardFunction} and \autoref{eq:rewardFunction2}, we present a general formula for the reward function.
We now discuss how to interpret a number of classical graph drawing algorithms with MARL based on such a formulation. 
For the FR and DCG layouts, the reward function for agent $v$ at time $t$ follows~\autoref{eq:rewardFunction}.
For the stress majorization, the reward function is given by~\autoref{eq:rewardFunction2}. 

\para{FR layout.}
The layout algorithm proposed by Fruchterman and Reingold~\cite{FruchtermanReingold1991} can be adjusted to our model by rewriting the total force on node $v$ at time $t$ as $F^v(t) = \sum_{(u,v) \in E}f^t_a(u,v) + \sum_{u \in V}f^t_r(u,v)$, for attractive forces $f_a^t$ and repulsive forces $f_r^t$ at time $t$. 
$F^v(t)$ can be written as 
\begin{equation}
F^v(t) = \sum_{(u,v) \in E}{{\frac{{d_t(u,v)}^2}{k}}} + \sum_{u\in V} \frac{-k^2}{{d_t(u,v)}}, 
\label{eq:fr-force}
\end{equation}
where $d_t(u,v)$ is the distance between $u$ and $v$ in the layout at time $t$. 

\para{DGC layout.}
Similarly, for the algorithm proposed by Dogrusoz \etal~\cite{DogrusozGiralCetintas2009}, the total force on node $v$ at time $t$ is written as  
\begin{equation}
F^v(t) = \sum_{(u,v) \in E}{\frac{{(\lambda - d_t(u,v))}^2}{\zeta}} + \sum_{u\in V} {\frac{\mu}{d_t(u,v)^2}}. 
\label{eq:dgc-force}
\end{equation}

\para{Stress majorization.}
The optimization of the stress function used by Gansher \etal~\cite{GansnerKorenNorth2005} can be formulated in two ways via MARL: optimizing the local stress on each agent, or optimizing the global stress via the action of an agent. 
To optimize the local stress on each agent, we define the \emph{local stress} for agent $v$  at time $t$ as
\begin{equation}
E^v(t) = \sum_{u \in N(v,p)} w_{u,v} (d_t(u,v) - d_t'(u,v))^2, 
\label{eq:localStress}
\end{equation}
where $N(v,p)$ denotes the nodes that are in the \emph{$p$-hop neighborhood} of $v$. 
That is, $u \in N(v,p)$ is at most $p$ hops away from $v$.

Alternatively, we also provide the \emph{global stress} function as the objective for each agent and rewrite~\autoref{eq:localStress} as
\begin{equation}
E^v(t) = \sum_{u < v} w_{u,v} (d_t(u,v) - d_t'(u,v))^2. 
\label{eq:globalStress}
\end{equation}
The reward function for agent $v$ at time $t$ is
$\mathcal{R}^{v} = E^v(t-1) - E^v(t)$, where $E^v(t)$ is either the local or the global stress function.

\subsection{Derive New Algorithms with Aesthetic Criteria}
\label{sec:newlayout}

A main advantage of our MARL model for graph drawing is its generality, that is, the ability to create new layout algorithms by customizing, for each agent, its reward function, as well as its state and action spaces.
In the most general form, a reward function $\Rcal^v$ on an agent $v$ at time $t$ can be used not only to encode forces and energy on the node (as shown in~\autoref{sec:interpretation}), but also to incorporate various aesthetic criteria for graph layouts. 

On a highlevel, a quality measure $Q$ for a graph layout is defined to be a linear combination of aesthetic criteria,
\begin{equation}
Q = \sum_i \omega_i q_i, \quad \sum_i \omega_i = 1
\label{eq:Q}
\end{equation}
where $q_i$ is the $i$-th aesthetic criterion, $\omega_i$ is its coefficient. 
Potential aesthetic criteria include the number of edge crossings, the amount of node overlaps, the distance between non-adjacent nodes, edge length variance, and the cumulative deviation of edge angles from the ideal values. 
These aesthetic criteria have been widely used in the evolutionary graph layout algorithms (e.g.,~\cite{BrankeBucherSchmeck1996,SuarezSebagRodriguez1999,Masui1994,Tettamanzi1998}).
The coefficients could be determined empirically~\cite{Tettamanzi1998} or learned using genetic programming~\cite{Masui1994}, as $Q$ is related (but not equivalent) to the fitness of a graph layout. 

In this paper, we experiment with a few aesthetic criteria from~\cite{Tettamanzi1998} and~\cite{Masui1994} below. 
The main idea is to define a local quality measure $Q^v$ to be associated with each agent $v$, and subsequently to encode $Q^v$ in the reward function $\Rcal^v$. 

\para{Edge length variance.}
For a given graph layout, the \emph{edge length variance} measures the mean relative square error of edge lengths~\cite{Tettamanzi1998}, 
$\sigma = \frac{1}{|E|}\sum_{e \in E} {\left(\frac{\lVert e \rVert - L}{L}\right)}^2,$
where $\|e\|$ is the length of an edge and $L$ is the desired edge length.
$\sigma$ can be modified locally for a single agent $v$,
$\sigma^v = \frac{1}{|E_v|}\sum_{e \in E_v} {\left(\frac{\lVert e \rVert - L}{L}\right)}^2,$
where $E_v$ represents edges incident to node $v$.

\para{Edge angle deviation.}
The \emph{edge angle deviation} captures the cumulative square deviation $\Delta$ of edge angles from their ideal values~\cite{Tettamanzi1998}, 
$\Delta = \sum_{v \in V} \sum_{k=1}^{|E_v|}{ \left( \psi_k(v) - \frac{2\pi}{|E_v|}\right)}^2, $
where $E_v$ contains edges incident to $v$, and 
$\psi_k(v)$ are the angles between adjacent edges in $E^v$. 
For an agent $v$, we can derive its local edge angle deviation as
$\Delta^v = \sum_{k=1}^{|E_v|} {\left( \psi_k(v) - \frac{2\pi}{|E_v|}\right)}^2.$

\para{Node overlaps.} 
For a given graph layout, let $\varphi$ represents the total number of node overlaps.  
Let $u < v$ denote an ordering among distinct vertices $u, v \in V$. 
\begin{equation}
\varphi = \sum_{u \in V, v \in V, u < v}
\begin{cases}
    1,& \text{if  } u \text{ and } v \text{ overlap};\\
    0,& \text{otherwise}. 
\end{cases}
\label{eq:node-overlap}
\end{equation}
For an agent $v$, the number of node overlaps local to $v$, denoted as $\varphi^v$, is the number of nodes in $V$ that overlaps with $v$, 
\begin{equation}
\varphi^v = \sum_{u \in V, u \neq v}
\begin{cases}
    1,& \text{if $u$ overlaps with $v$;}\\
    0,& \text{otherwise.}
\end{cases}
\end{equation}

\para{Edge crossings.}
Let $\chi$ represent the total number of edge crossings. 
For agent $v$, the number of edge crossings local to $v$ is defined as
\begin{equation}
\chi^v = \sum_{e_i \in E_v} \sum_{e_j \in E}
\begin{cases}
    1,& \text{if $e_i$ and $e_j$ cross in their interior;}\\
    0,& \text{otherwise;}
\end{cases}
\end{equation}
where $E_v$ represents edges incident to node $v$.

\para{Minimum distance between nodes.}
The aesthetic criteria $\eta$ ensures that there is enough space between nodes~\cite{Masui1994}.
For agent $v$, the local version of $\eta$ is defined as
\begin{equation}
\eta^v = \sum_{u \in V}
\begin{cases}
    L - d(u,v), & \text{if } d(u,v) < L;\\
    0, & \text{otherwise};
\end{cases}
\end{equation}
where $L$ is the expected minimum distance between nodes.

\para{Local quality measure.}
Given the above aesthetic criteria, we derive a local quality measure for an agent $v$ as 
\begin{equation}
Q^v = \omega_1  \varphi^v + \omega_2  \chi^v + \omega_3  \eta^v + \omega_4  \sigma^v + \omega_5  \Delta^v. 
\label{eq:quality}
\end{equation}
We then define a reward function $\Rcal^v$, 
\begin{equation}
\Rcal^v(t) = Q^v(t-1) - Q^v(t), 
\label{eq:rewardFunction3}
\end{equation}
based upon $Q^v$ evaluated at time $t-1$ and $t$. 
Notice that \autoref{eq:rewardFunction3} takes a similar form as ~\autoref{eq:rewardFunction} and~\autoref{eq:rewardFunction2}. 

\subsection{Extensions to Incremental and Hybrid Layouts} 
Finally, our MARL model can be employed to interpret incremental graph layout algorithms. 
In incremental layouts, instead of computing a new layout for the entire graph, only a small part of the graph is rearranged, and the positions of the remaining nodes stay (roughly) the same~\cite{MiriyalaHornickTamassia1993,North1996}.
Incremental layouts are important for maintaining a mental map of the users, which helps them keep track of changes in the graph~\cite{PurchaseHogganGorg2007}.
Incremental layouts can also be used for online dynamic graphs~\cite{CrnovrsaninChuMa2015,BeckBurchDiehl2017}. 

Our MARL model can be modified in various ways to act as an incremental layout algorithm.
One obvious approach would be to lock a node (or nodes) that we do not want to move by limiting the action space of its agent (to only contain the ``stay" action). 
Another approach is to modify the reward function, that is, nodes are still allowed to move, but any action other than the ``stay" action has a lower reward. 
Restricting the state and action spaces of our MARL model will allow us to create an incremental version of most classical graph drawing algorithms; see the supplementary video for a demo.

Finally, an additional advantage of our MARL model is that the reward function is highly customizable. 
For instance, a hybrid reward function can be created by taking a linear combination of force-based, energy-based, and/or aesthetic-based rewards, which is difficult to do with classic layout algorithms. As an example, we can create a new layout algorithm that is a hybrid of a FR layout and stress majorization: 
\begin{equation}
    \Rcal^v(t) = \beta (F^v(t-1) - F^v(t)) + (1-\beta) (E^v(t-1) - E^v(t)), 
    \label{eq:hybrid}
\end{equation}
where $\beta$ is the weight, $F^v(t)$ and $E^v(t)$ are given in~\autoref{eq:fr-force} and~\autoref{eq:localStress}, respectively.

\section{Algorithm and Implementation} 
\label{sec:implementation}

The proposed MARL framework for graph drawing is sketched in~\autoref{algorithm:marl}. 
This general framework is applicable to the MARL versions for DCG layout, FR layout, and stress majorization
as well as to MARL layouts with hybrid and customized rewards. 

\begin{algorithm}[ht]
	\SetKwInOut{Input}{input}
	\SetKwInOut{Output}{output}
	\SetKwData{globalLayout}{globalLayout}
	\SetKwData{Gi}{$G_i$}
	\SetKwData{intersectionGraph}{intersectionGraph}
	\SetKwData{layout}{layout}
	\SetKwData{curObjectiveValue}{$\Rcal^v$}
	\SetKwData{action}{$a^v$}
	\SetKwData{reward}{$\Rcal^v$}
	\SetKwFunction{initPositionsRandomly}{initialize the location of $v$ randomly}
	\SetKwFunction{initAgentsForEachNode}{create and assign an agent to node $v$}
	\SetKwFunction{rewardFunction}{calculateReward}
	\SetKwFunction{getNextAction}{getNextAction}
	\SetKwFunction{updateNodePosition}{moveNode}
	\SetKwFunction{updateRewardTable}{updateRewardTable}
	\SetKwFunction{staticLayouter}{staticLayouter}
	\SetKwFunction{drawGraph}{drawGraph}
	\SetKwFunction{modifiedStaticLayouter}{modifiedStaticLayouter}
	\SetKwFunction{setPositionsOfCommonNodes}{setPositionsOfCommonNodes}
	\Input{A graph $G(E,V)$, a reward function $\Rcal$}
	\Output{The position of each node $v \in V$}
	\BlankLine
	\tcp{initialization}
	\For{$v \in V$}{
	\initAgentsForEachNode\;
    	\initPositionsRandomly\;
	}
	\BlankLine
	\tcp{iterative layout procedure}
	\While{not converged}{
    	\For{$v \in V$}{
	    \tcp{obtain reward for the agent}
    	    \curObjectiveValue$\leftarrow$ \rewardFunction{$v$}\; 
	    \tcp{obtain the next action for the agent}
    	    \action$\leftarrow$ \getNextAction{$v$}\;
	    \tcp{update the location of node}
    	    $v$ $\leftarrow$ \updateNodePosition{$v$, \action}\;
	    \tcp{update the reward of the agent}
    	    \reward$\leftarrow$ \curObjectiveValue\ - \rewardFunction{$v$}\;
    	    \updateRewardTable{$v$, \reward}\;
    	}
	}
	\caption{MARL Graph Layout}
	\label{algorithm:marl}
\end{algorithm}

The $\mathsf{calculateReward}$ function evaluates the force-based, energy-based, or customized reward for a given agent. 
The $\mathsf{getNextAction}$ function returns the next action based on the greedy policy of an agent.
Once the next action is obtained, the node controlled by an agent is moved in the direction of the given action via $\mathsf{moveNode}$. 
The $\mathsf{updateRewardTable}$ updates the Q-function values based on~\autoref{eq:qFunction}.

The proposed MARL framework API is implemented as a \textsf{Cytoscape.js}~\cite{FranzLopesHuck2016} extension in JavaScript in combination with \textsf{REINFORCEjs}\footnote{{https://github.com/karpathy/reinforcejs}}, a reinforcement learning library.  
For reproducibility, parameter settings for all graph drawing algorithms described in this paper is detailed in~\autoref{sec:implementation-details}.

\section{A Visual Demo via {\tool}}
\label{sec:interface}

\begin{figure}[!ht]
\centering
\includegraphics[width=1.0\columnwidth]{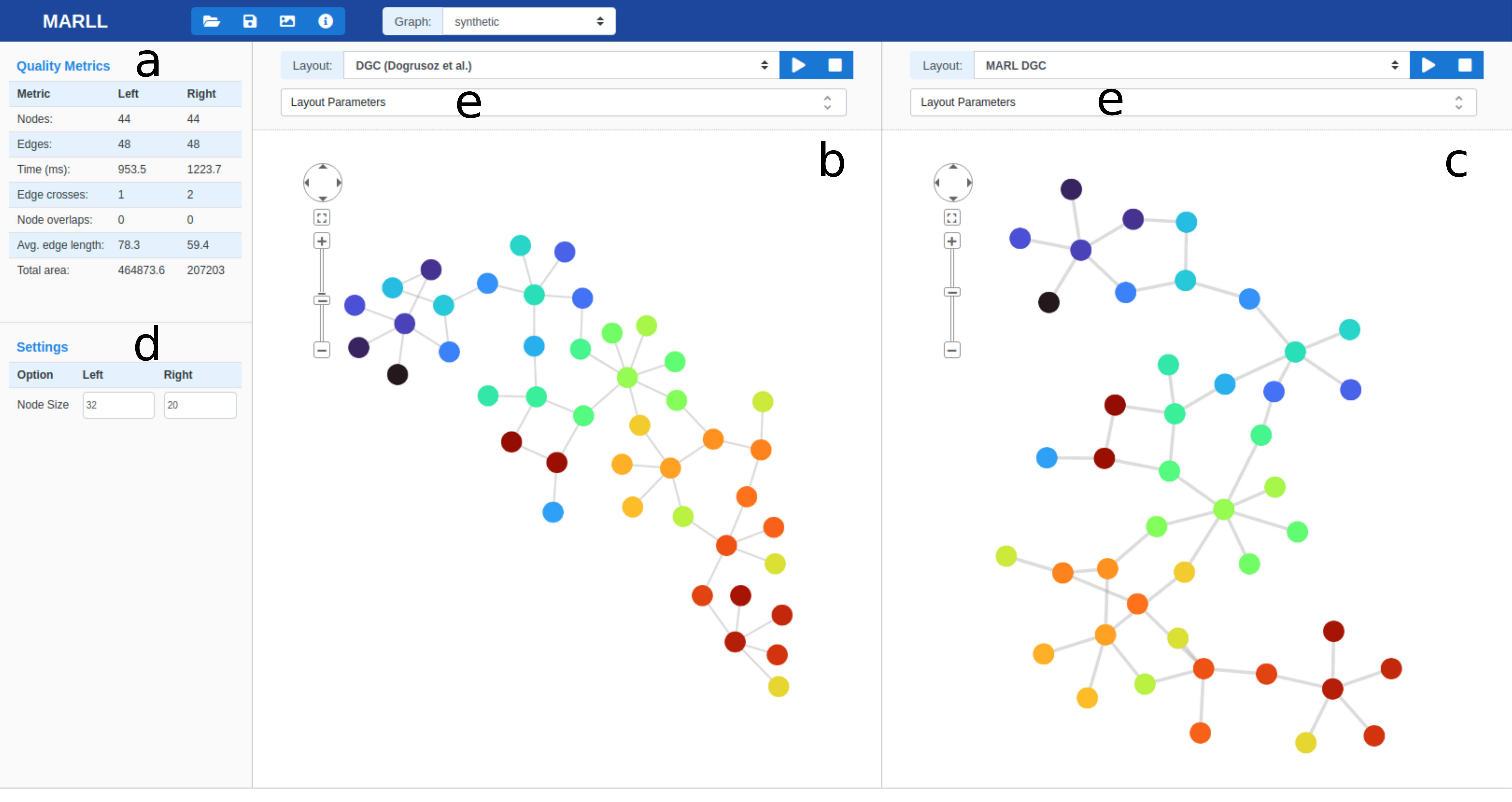}
\vspace{-4mm}
\caption{The interactive user interface of {\tool}.}
\label{fig:interface}
\vspace{-2mm}
\end{figure}

In addition to an API, we also provide an open-source, interactive tool called {\tool}.  
{\tool} demonstrates the capabilities of our MARL algorithm, evaluates and compares the resulting graph layouts against classic layout algorithms. 
It is built with HTML/CSS/Javascript stack with \textsf{Cytoscape.js}, \textsf{D3.js} and \textsf{JQuery} Javascript libraries. 
{\tool} allows users to experiment with the MARL framework by providing their own graph files or choosing among existing sample graphs. 
Its user interface is shown in~\autoref{fig:interface}. 

The user interface consists of three main panels.
Panel (a) contains statistics and quality metrics used to evaluate current graph layouts quantitatively. 
To facilitate layout comparisons, the interface provides two adjacent linked views -- panel (b) and panel (c) -- that visualize a graph with two different layout algorithms side-by-side.  
Elements (nodes and edges) highlighted in panel (b) will be highlighted in panel (c), and vice versa. 
Panel (d) allows users to change the sizes of nodes in either view.
Users can also use panel (e) to modify the parameters of the selected layout algorithms. 
The nodes are colored based on their indices using Turbo, an improved  rainbow~\cite{Mikhailov2019} colormap, which helps identify corresponding nodes between the two views.

\section{Experiments and Evaluation}
\label{sec:evaluation}
In this section, we perform quantitative and qualitative evaluations together with runtime analysis. 
We demonstrate via experiments that MARL graph layout algorithms produce results that are aesthetically comparable to those of the classical layout algorithms.  
At the same time, MARL layout algorithms are easily generalizable and provide a unifying framework for a set of classic layout algorithms. 

\subsection{Datasets}
For our experiments, we use 10 graphs with varying types and sizes.
The graphs, descriptions, and sources are given in~\autoref{tab:dataset}. 
 
\begin{table}[!ht]
    \centering
    \small
    \caption{The graphs used in our experiments.} 
    \begin{tabular}{lllccc}
		\toprule[1pt]\midrule[0.3pt]
         \textbf{Graph} & \textbf{Name} & \textbf{Type} &$\mathbf{|V|}$ & $\mathbf{|E|}$ & \textbf{Source}\\ \midrule[0.3pt]
         $G_1$ & karate & social network& 34 & 78 & \cite{Kunegis2013,Zachary1977}\\
         $G_2$ & bcspwr01 & power system& 39 & 46 & \cite{DavisHu2011}\\
         $G_3$ & synthetic & synthetic & 44 & 48 & \cite{DogrusozGiralCetintas2009}\\
         $G_4$ & lesmiserables & social network& 77 & 254 & \cite{Kunegis2013,Knuth1993}\\
         $G_5$ & GD99\_c & graph drawing contest& 105 & 149 & \cite{DavisHu2011}\\
         $G_6$ & bcspwr03 & power system& 118 & 179 & \cite{DavisHu2011}\\
         $G_7$ & rajat05 & circuit simulation & 301 & 952 & \cite{DavisHu2011}\\
         $G_8$ & GD00\_c & graph drawing contest& 638 & 1031 & \cite{DavisHu2011}\\
         $G_9$ & crime\_moreno & social network & 829 & 1473 & \cite{DavisHu2011}\\
         $G_{10}$ & maayan-faa & flight network & 1226 & 2613 & \cite{RossiAhmed2015}\\
 		\midrule[0.3pt]\bottomrule[1pt]
    \end{tabular}
    \label{tab:dataset}
\end{table}

\subsection{Parameter Configurations for Evaluation Purposes}
\label{sec:para}

We implement and evaluate classic graph layout algorithms in comparison with their MARL counterparts. 

Specifically, we include force-directed layouts such as \textbf{FR} layout by Fruchterman and Reingold~\cite{FruchtermanReingold1991} and \textbf{DGC} layout by Dogrusoz \etal~\cite{DogrusozGiralCetintas2009}; their MARL versions are denoted as \textbf{MARL FR} and \textbf{MARL DGC}, respectively. 
The \textbf{MARL Custom} layout is based on local quality measure in \autoref{eq:rewardFunction3}. 
For a reasonable comparison, these layout algorithms are ran until convergence with the same convergence criteria, that is, with displacement rate $\Delta A = 2$ (pixels). 
We also include stress majorization (\textbf{SM}) by Gansner \etal~\cite{GansnerKorenNorth2005}, whose MARL counterparts are denoted as  \textbf{MARL Global Stress} and \textbf{MARL Local Stress}, respectively, based on~\autoref{sec:method}. 
Their shared convergence criterion is the stress ratio $\delta E = 0.0001$. 
Finally, we create a hybrid reward function in the MARL setting (\textbf{MARL Hybrid}), where the algorithm converges based on both displacement rate and stress ratio (whichever comes first). 

For each of the layout algorithm, the node positions are initialized randomly.
We run each layout algorithm $100$ times and report the average evaluation measures (such as aesthetic criteria and runtime). 
The evaluation is conducted on a laptop machine with an Intel i7-9850H processor (6 cores at 2.6 GHz) CPU and 64 GB of RAM. 

\subsection{Quantitative Evaluation} 
\label{sec:quantitativeEvalution}

To evaluate the layouts quantitatively, we employ a list of aesthetic criteria   from the literature (e.g.~\cite{Tettamanzi1998, Purchase2002,PurchasePilcherPlimmer2012}), namely, minimizing the number of node overlaps, reducing the number of edge crossings~\cite{Purchase2002}, maximizing the minimum angle between edges leaving a node~\cite{Purchase2002} (similar to the the \emph{edge angle deviation}~\cite{Tettamanzi1998}), and minimizing the mean relative square error of edge lengths~\cite{Tettamanzi1998} (the \emph{edge length variance}); see~\autoref{sec:newlayout} for their definitions. 

For comparative purposes, the measurement for each aesthetic criteria is normalized to be a real number in $[0,1]$ following the metrics proposed by Purchase~\cite{Purchase2002}, where a higher value indicates a better layout aesthetically. 
Such a normalization ensures that ``the metric value does not depend on the nature of the underlying graph"~\cite{Purchase2002}.

\para{Normalized edge crossings (NC).} 
Recall $|V| = n$ and $|E| = m$. 
Let $\chi$ denote the number of edge crossings in a layout. 
Let $\cmx$ denote an approximation for the upper bound on the number of edge crossings~\cite{Purchase2002}, 
\begin{equation}
\cmx = \frac{m(m-1)}{2} - \frac{1}{2}\sum_{i=1}^{n} \deg(v_i)(\deg(v_i) - 1).
\label{eq:cmx}
\end{equation}
In~\autoref{eq:cmx}, the first component is the maximum possible edge crossings, and the second component is the total number of  impossible crossings (since adjacent edges cannot cross). 

The normalized number of edge crossings (abbreviated as \textbf{NC}) is based on the \emph{edge crossings aesthetic metric} of Purchase~\cite{Purchase2002}, 
\begin{equation}
N_c = 1- 
\begin{cases}
    \frac{\chi}{\cmx},& \text{if } \cmx > 0;\\
    0,& \text{otherwise}.
\end{cases}
\label{eq:nc}
\end{equation}

\para{Normalized node overlaps (NO).}
Let $\binom{n}{2} = n(n-1)/2$ denote the maximum number of node overlaps in a layout.  
Recall $\varphi$ is the total number of node overlaps, see~\autoref{eq:node-overlap}. 
We define the \emph{normalized number of node overlaps} (abbreviated  as \textbf{NO}) as 
\begin{equation}
N_o = 1 - \frac{\varphi}{{n(n-1)/2}}.
\label{eq:no}
\end{equation}

\para{Normalized edge lengths (NE).}
The normalized edge lengths (abbreviated as \textbf{NE}) measure how uniform the edge lengths are.
We define NE as
$$
N_e = \frac{1}{1 + \sigma}
$$
where $\sigma$ is the mean relative square error of edge lengths (the edge length variance).

\para{Normalized edge angles (NA).} 
The normalized edge angle (abbreviated as \textbf{NA}) is based on the \emph{minimum angle aesthetic metric}~\cite{Purchase2002}. 
It measures how well the edge angles are distributed surrounding a given node. 
For instance, for a degree-4 node, the ideal angle is 90 degree.
The normalized edge angle is defined as 
\begin{equation}
N_a = 1 - \frac{1}{n} \sum_{i=1}^{n} \left| \frac{\theta^*_i - \theta_{i}}{\theta^*_i} \right|, 
\label{eq:na}
\end{equation}
where $\theta^*_i$ is the ideal (maximal) minimum angle at the $i$-th node,
$
\theta^*_i = {360^\circ}/{\deg(v_i)};
$
and $\theta_{i}$ is the actual minimum angle between the incident edges at the $i$-th node~\cite{Purchase2002}.

\para{Results.} 
We use these four metrics (\textbf{NC, NO, NE,} and \textbf{NA}) to perform quantitative evaluations.
The results are given in~\autoref{tab:fd-sm-evaluation-results} and~\autoref{tab:marl-custom-reward-evaluation-results}, respectively.   

In~\autoref{tab:fd-sm-evaluation-results} , \textbf{\RNC, \RNO, \RNE,} and \textbf{\RNA} represent the ratio of measurements for the MARL version to the classical version, respectively.
For example, $\RNC$ represents the ratio of the number of edge crossings for MARL version to the number of edge crossings for the classical version. 
The higher the ratios are, the more aesthetically comparable the results are between the MARL layouts and the classical layouts.

\begin{table*}[htp]
\caption{Quantitative evaluation results for classical force-directed layouts, stress majorization, and their MARL versions.}
\label{tab:fd-sm-evaluation-results}
\small

\aboverulesep=0ex
\belowrulesep=0ex

	\begin{tabu} to \textwidth {X|*{4}X|XXXX >{\bfseries}X >{\bfseries}X >{\bfseries}X >{\bfseries}X| *{4}X|XXXX >{\bfseries}X >{\bfseries}X >{\bfseries}X >{\bfseries}X}
		\toprule[1pt]\midrule[0.3pt]
    \multicolumn{1}{c|}{\textbf{Graph}} & \multicolumn{4}{c|}{\textbf{DGC}} & \multicolumn{8}{c|}{\textbf{MARL DGC}} & \multicolumn{4}{c|}{\textbf{FR}} & \multicolumn{8}{c}{\textbf{MARL FR}} \\
    \cmidrule(lr){2-5} \cmidrule(lr){6-13} \cmidrule(lr){14-17} \cmidrule(lr){18-25}
    \multicolumn{1}{c|}{\textit{}} & {NC} & {NO} & {NE} & {NA} & {NC} & {NO} & {NE} & {NA} & $\RNC$ & $\RNO$ & $\RNE$ & $\RNA$ & {NC} & {NO} & {NE} & {NA} & {NC} & {NO} & {NE} & {NA} & $\RNC$ & $\RNO$ & $\RNE$ & $\RNA$ \\ \midrule[0.3pt]
$G_1$ & 0.96 & 1 & 0.91 & 0.24 & 0.91 & 0.99 & 0.85 & 0.22 & 0.95 & 0.99 & 0.93 & 0.91 & 0.96 & 1 & 0.93 & 0.25 & 0.95 & 0.97 & 0.86 & 0.23 & 0.99 & 0.97 & 0.93 & 0.91 \\
$G_2$ & 1    & 1 & 1    & 0.73 & 0.95 & 1    & 0.93 & 0.65 & 0.95 & 1    & 0.93 & 0.89 & 1    & 1 & 1    & 0.73 & 0.99 & 0.99 & 0.91 & 0.64 & 0.99 & 0.99 & 0.91 & 0.88 \\
$G_3$ & 1    & 1 & 1    & 0.83 & 0.96 & 1    & 0.96 & 0.77 & 0.96 & 1    & 0.96 & 0.93 & 1    & 1 & 1    & 0.81 & 0.99 & 0.99 & 0.9  & 0.73 & 0.99 & 0.99 & 0.9  & 0.9 \\
$G_4$ & 0.96 & 1 & 0.86 & 0.35 & 0.92 & 0.99 & 0.8  & 0.33 & 0.96 & 0.99 & 0.92 & 0.93 & 0.97 & 1 & 0.89 & 0.37 & 0.95 & 0.98 & 0.79 & 0.34 & 0.99 & 0.98 & 0.89 & 0.92 \\
$G_5$ & 1    & 1 & 0.99 & 0.72 & 0.96 & 1    & 0.92 & 0.64 & 0.96 & 1    & 0.92 & 0.88 & 1    & 1 & 0.99 & 0.71 & 0.98 & 0.99 & 0.89 & 0.62 & 0.98 & 0.99 & 0.9  & 0.87 \\
$G_6$ & 1    & 1 & 0.98 & 0.54 & 0.95 & 0.99 & 0.91 & 0.5  & 0.95 & 0.99 & 0.94 & 0.93 & 1    & 1 & 0.97 & 0.54 & 0.97 & 0.99 & 0.87 & 0.48 & 0.98 & 0.99 & 0.89 & 0.89 \\
$G_7$ & 0.99 & 1 & 0.9  & 0.34 & 0.93 & 0.99 & 0.82 & 0.32 & 0.93 & 0.99 & 0.91 & 0.93 & 0.99 & 1 & 0.9  & 0.34 & 0.97 & 0.99 & 0.84 & 0.31 & 0.98 & 0.99 & 0.92 & 0.93 \\
$G_8$ & 0.99 & 1 & 0.88 & 0.61 & 0.89 & 0.99 & 0.81 & 0.55 & 0.9  & 0.99 & 0.92 & 0.89 & 0.99 & 1 & 0.88 & 0.62 & 0.94 & 0.99 & 0.86 & 0.56 & 0.95 & 0.99 & 0.98 & 0.9 \\
$G_9$ & 0.97 & 1 & 0.88 & 0.5  & 0.87 & 0.99 & 0.81 & 0.44 & 0.9  & 1    & 0.92 & 0.88 & 0.97 & 1 & 0.89 & 0.5  & 0.93 & 0.99 & 0.87 & 0.45 & 0.95 & 0.99 & 0.98 & 0.9 \\
$G_{10}$ & 0.99 & 1 & 0.87 & 0.43 & 0.85 & 1    & 0.84 & 0.37 & 0.86 & 1    & 0.97 & 0.85 & 0.99 & 1 & 0.87 & 0.44 & 0.92 & 0.99 & 0.84 & 0.39 & 0.93 & 0.99 & 0.96 & 0.89 \\
 \midrule[0.3pt]\bottomrule[1pt]
\end{tabu}

\vspace{1em}
	\begin{tabu} to \textwidth {X| *{4}X| XXXX >{\bfseries}X >{\bfseries}X >{\bfseries}X >{\bfseries}X| *{4}X >{\bfseries}X >{\bfseries} X >{\bfseries}X >{\bfseries}X |XXXX}
		\toprule[1pt]\midrule[0.3pt]
    \multicolumn{1}{c|}{\textbf{Graph}} & \multicolumn{4}{c|}{\textbf{SM}} & \multicolumn{8}{c|}{\textbf{MARL Local Stress}} & \multicolumn{8}{c|}{\textbf{MARL Global Stress}} & \multicolumn{4}{c}{\textbf{MARL Hybrid}} \\
    \cmidrule(lr){2-5} \cmidrule(lr){6-13} \cmidrule(lr){14-21} \cmidrule(lr){22-25}
    \multicolumn{1}{c|}{\textit{}} & {NC} & {NO} & {NE} & {NA} & {NC} & {NO} & {NE} & {NA} & $\RNC$ & $\RNO$ & $\RNE$ & $\RNA$ & {NC} & {NO} & {NE} & {NA} & $\RNC$ & $\RNO$ & $\RNE$ & $\RNA$ & {NC} & {NO} & {NE} & {NA} \\ \midrule[0.3pt]
$G_1$ & 0.96 & 1 & 0.93 & 0.25 & 0.95 & 0.99 & 0.86 & 0.22 & 0.99 & 0.99 & 0.92 & 0.88 & 0.95 & 0.99 & 0.86 & 0.22 & 0.99 & 0.99 & 0.92 & 0.87 & 0.95 & 0.99 & 0.86 & 0.22 \\
$G_2$ & 1    & 1 & 1    & 0.73 & 0.99 & 1    & 0.91 & 0.67 & 0.99 & 1    & 0.91 & 0.92 & 0.99 & 1    & 0.9  & 0.66 & 0.99 & 1    & 0.91 & 0.91 & 0.99 & 1 & 0.91 & 0.67 \\
$G_3$ & 1    & 1 & 1    & 0.81 & 0.99 & 1    & 0.92 & 0.77 & 0.99 & 1    & 0.92 & 0.94 & 0.99 & 1    & 0.92 & 0.76 & 0.99 & 1    & 0.93 & 0.94 & 0.99 & 1 & 0.92 & 0.76 \\
$G_4$ & 0.97 & 1 & 0.89 & 0.37 & 0.95 & 0.99 & 0.83 & 0.32 & 0.98 & 0.99 & 0.93 & 0.87 & 0.95 & 0.99 & 0.83 & 0.32 & 0.98 & 0.99 & 0.93 & 0.88 & 0.94 & 0.99 & 0.84 & 0.32 \\
$G_5$ & 1    & 1 & 0.99 & 0.71 & 0.99 & 1    & 0.89 & 0.65 & 0.99 & 1    & 0.9  & 0.92 & 0.99 & 1    & 0.91 & 0.65 & 0.99 & 1    & 0.92 & 0.91 & 0.99 & 1 & 0.9 & 0.65 \\
$G_6$ & 1    & 1 & 0.97 & 0.54 & 0.98 & 1    & 0.9  & 0.49 & 0.99 & 1    & 0.93 & 0.91 & 0.99 & 1    & 0.88 & 0.48 & 0.99 & 1    & 0.9  & 0.9  & 0.98 & 1 & 0.9 & 0.5 \\
$G_7$ & 0.99 & 1 & 0.9  & 0.34 & 0.99 & 1    & 0.81 & 0.3  & 1    & 1    & 0.9  & 0.88 & 0.99 & 1    & 0.82 & 0.29 & 0.99 & 1    & 0.9  & 0.87 & 0.99 & 1 & 0.82 & 0.3 \\
$G_8$ & 0.99 & 1 & 0.88 & 0.62 & 0.98 & 1    & 0.79 & 0.58 & 0.99 & 1    & 0.89 & 0.93 & 0.98 & 1    & 0.79 & 0.57 & 0.99 & 1    & 0.89 & 0.93 & 0.98 & 1 & 0.79 & 0.58 \\
$G_9$ & 0.97 & 1 & 0.89 & 0.5  & 0.96 & 1    & 0.8  & 0.46 & 0.99 & 1    & 0.9  & 0.91 & 0.96 & 1    & 0.81 & 0.46 & 0.99 & 1    & 0.92 & 0.91 & 0.96 & 1 & 0.78 & 0.46 \\
$G_1$ & 0.99 & 1 & 0.87 & 0.44 & 0.98 & 1    & 0.76 & 0.37 & 0.99 & 1    & 0.87 & 0.86 & 0.98 & 1    & 0.77 & 0.37 & 0.99 & 1    & 0.88 & 0.86 & 0.98 & 1 & 0.77 & 0.37 \\
 \midrule[0.3pt]\bottomrule[1pt]
\end{tabu}

\end{table*}

In~\autoref{tab:marl-custom-reward-evaluation-results}, we give quantitative evaluation results using a local quality measure as the custom reward function (\autoref{eq:rewardFunction3}). 
The results indicate that our MARL framework with a custom reward function produces aesthetically comparable layouts in comparison with other algorithms, shown in~\autoref{tab:fd-sm-evaluation-results}.  

\begin{table}[!htbp]
\caption{Quantitative evaluation results for MARL layouts based on custom reward function.}
\label{tab:marl-custom-reward-evaluation-results}
	\begin{tabu} to \columnwidth {XXXXX}
		\toprule[1pt]\midrule[0.3pt]
    \multicolumn{1}{c}{\textbf{Graph}} & \multicolumn{4}{c}{\textbf{MARL Custom Reward}}\\
    \cmidrule(lr){2-5}
    \multicolumn{1}{c}{\textit{}} & {NC} & {NO} & {NE} & {NA}\\ \midrule[0.3pt]
$G_1$& 0.931&0.989&0.875&0.243 \\
$G_2$& 0.963&0.992&0.896&0.581 \\
$G_3$& 0.96&0.993&0.859&0.687 \\
$G_4$& 0.931&0.993&0.779&0.357 \\
$G_5$& 0.951&0.994&0.849&0.591 \\
$G_6$& 0.936&0.993&0.825&0.456 \\
$G_7$& 0.962&0.996&0.778&0.29 \\
$G_8$& 0.91&0.996&0.767&0.559 \\
$G_9$& 0.913&0.996&0.753&0.459 \\
$G_{10}$& 0.937&0.997&0.774&0.397 \\
\midrule[0.3pt]\bottomrule[1pt]
\end{tabu}
\end{table}

The quantitative evaluations show that the MARL layout algorithms produce results very similar to those of their classical counterparts in terms of the four quality metrics, independent of the graph size.
The two MARL layout algorithms based on the custom and hybrid reward functions do not have a classical version for a direct comparison.
However, their quality metrics are close to those for classical layout algorithms.

\subsection{Qualitative Evaluation}
\begin{figure*}
    \centering
    \includegraphics[width=0.99\textwidth]{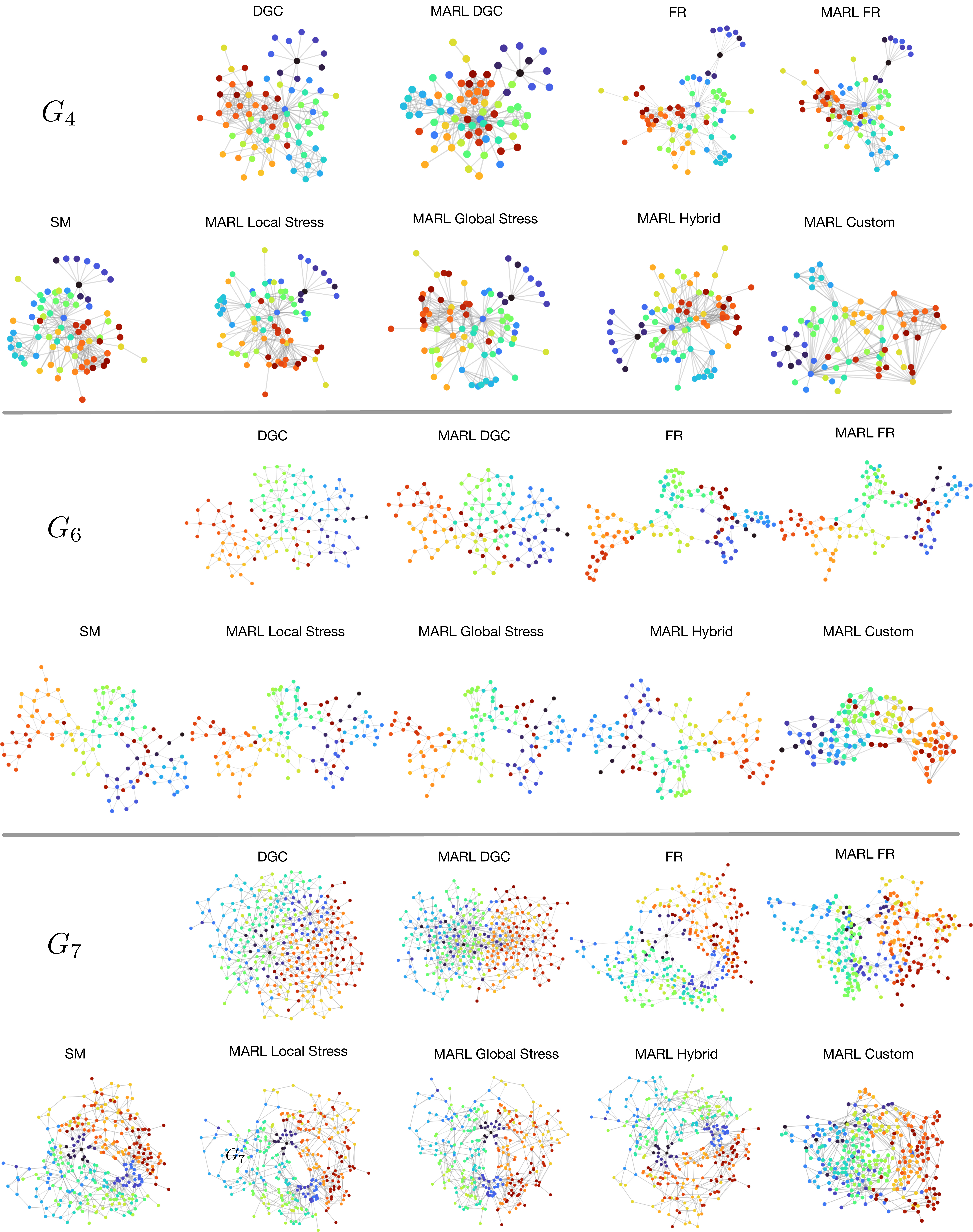}
    \caption{Qualitative results for $Q_4$ (top), $G_6$ (middle) and $Q_7$ (bottom) for DGC layout, FR layout, stress majorization (SM), and their MARL versions; as well as MARL hybrid and MARL custom versions. }
    \label{fig:evaluation-G4-G7}
\end{figure*}

We also perform a qualitative evaluation to demonstrate that the MARL framework produces results aesthetically comparable to those of the classical layout algorithms.
We show examples $G_4$, $G_6$, and $G_7$ in~\autoref{fig:evaluation-G4-G7}, where we compare the layouts obtained by classical algorithms (DGC, FR, and stress majorization) with those obtained by their MARL versions. 
We also include layouts obtained via a MARL hybrid and MARL custom. 
For each graph, nodes with the same color correspond to one another. 
Due to space constraints, additional layouts for all graphs can be found in~\autoref{sec:evaluation-details}. 

Based on our qualitative evaluation, the classical layout algorithms and the MARL versions produce graph drawings that are visually alike.
For some of the graphs, the results are visually indistinguishable. 
The layouts of the MARL hybrid layout are as good as the layouts from the local and global stress MARL algorithm.
Since the MARL layout with a custom reward function does not have a classical counterpart, its outputs are not expected to be visually similar to others.

\subsection{Runtime Analysis}
Finally, we compare the runtime of classic layout algorithms until convergence against their corresponding MARL versions; see~\autoref{tab:fd-runtime-analysis} for force-directed layout algorithms and~\autoref{tab:sm-runtime-analysis} for stress majorization. 
We follow the parameter configurations detailed in~\autoref{sec:para}. 
When applicable, we report the ratio of runtime of MARL layouts to their classical counterparts under the same convergence criteria.
All runtimes are reported in milliseconds.

\begin{table}[!ht]
    \caption{Runtimes in milliseconds for classical and MARL force-directed layouts.}
    \label{tab:fd-runtime-analysis}
	\begin{tabu} to \columnwidth {XXXXXXX}
		\toprule[1pt]\midrule[0.3pt]
    \multicolumn{1}{c}{\textbf{Graph}} & \multicolumn{6}{c}{\textbf{Runtime of Force Directed Layouts}}\\
    \cmidrule(lr){2-7}
    \multicolumn{1}{c}{\textit{}} & {DGC} & {MARL DGC} & {Ratio DGC} & {FR} & {MARL FR} & {Ratio FR}\\ \midrule[0.3pt]
$G_1$ & 72&438&6.05&140&352&2.51 \\
$G_2$ & 74&458&6.17&157&397&2.53 \\
$G_3$ & 85&582&6.85&304&481&1.58 \\
$G_4$ & 285&1744&6.11&606&1011&1.67 \\
$G_5$ & 276&1953&7.08&705&886&1.26 \\
$G_6$ & 328&2563&7.81&884&1008&1.14 \\
$G_7$ & 3598&16810&4.67&4030&5052&1.25 \\
$G_8$ & 15383&68801&4.47&13688&18597&1.36 \\
$G_9$ & 34347&82755&2.41&26890&32378&1.2 \\
$G_{10}$ & 70100&143892&2.05&54684&68022&1.24 \\
\midrule[0.3pt]\bottomrule[1pt]
\end{tabu}
\end{table}

\begin{table}[!htbp]
\caption{Runtimes in milliseconds for classical stress majorization algorithm and its MARL variants.}
\label{tab:sm-runtime-analysis}
	\begin{tabu} to \columnwidth {XXXXXX}
		\toprule[1pt]\midrule[0.3pt]
    \multicolumn{1}{c}{\textbf{Graph}} & \multicolumn{5}{c}{\textbf{Runtime of Stress Majorization Layouts}}\\
    \cmidrule(lr){2-6}
    \multicolumn{1}{c}{\textit{}} & {SM} & {MARL Local Stress} & Ratio Local Stress & MARL Global Stress & Radio Global Stress \\ \midrule[0.3pt]
$G_1$ & 17&126&7.46&135&7.99 \\
$G_2$ & 20&177&8.88&187&9.36 \\
$G_3$ & 25&249&9.95&257&10.27 \\
$G_4$ & 64&398&6.42&386&6.23 \\
$G_5$ & 50&550&10.96&605&12.07 \\
$G_6$ & 65&630&9.74&633&9.79 \\
$G_7$ & 544&5154&9.48&5235&9.63 \\
$G_8$ & 6700&25696&3.84&26050&3.89 \\
$G_9$ & 10884&42274&3.88&42730&3.93 \\
$G_{10}$ & 36360&118486&3.26&121637&3.35 \\
\midrule[0.3pt]\bottomrule[1pt]
\end{tabu}
\end{table}

The runtimes for the MARL layouts with custom (\autoref{eq:rewardFunction3}) and hybrid (\autoref{eq:hybrid}) reward functions are provided in~\autoref{tab:hybrid-custom-runtime-analysis}.

\begin{table}[!htbp]
\caption{Runtimes in milliseconds for the MARL layouts with custom and hybrid reward functions.}
\label{tab:hybrid-custom-runtime-analysis}
	\begin{tabu} to \columnwidth {XXX}
		\toprule[1pt]\midrule[0.3pt]
    \multicolumn{1}{c}{\textbf{Graph}} & \multicolumn{2}{c}{\textbf{Runtime of Hybrid and Custom}}\\
    \cmidrule(lr){2-3}
    \multicolumn{1}{c}{\textit{}} & \textit{MARL Hybrid} & \textit{MARL Custom}
    \\ \midrule[0.3pt]
$G_1$ & 128&394 \\
$G_2$ & 186&454 \\
$G_3$ & 269&474 \\
$G_4$ & 430&1552 \\
$G_5$ & 702&1110 \\
$G_6$ & 746&1725 \\
$G_7$ & 6304&12112 \\
$G_8$ & 30764&126504 \\
$G_8$ & 45681&232724 \\
$G_{10}$ & 138034&752484 \\
\midrule[0.3pt]\bottomrule[1pt]
\end{tabu}
\end{table}

In comparison to the classical algorithms, the runtime overhead for MARL versions is very large for smaller sized graphs, especially for the MARL layout based on~\cite{DogrusozGiralCetintas2009} and~\cite{GansnerKorenNorth2005}.
However, as the graph size increases, the difference between the run-time of classical and their MARL versions decreases. 
The MARL Hybrid has a computation time similar to that of the MARL layouts based on local stress and global stress.
Although the MARL Custom does not have a classical version for comparison, its generally has the slowest computation time among all MARL layouts.

\section{Discussion}
\label{sec:discussions}

In this paper, we propose a novel MARL-based approach to graph drawing. 
Our approach interprets several classical force-directed and energy-based graph drawing algorithms with MARL. 
Additionally, it enables derivation of new layout algorithms by providing custom and hybrid reward functions to the agents of the system.

The main attraction for a MARL formulation is that it is flexible, and can be adapted to existing and new graph drawing algorithms. 
By modeling with MARL, we provide a unifying framework for a number of classical layout algorithms, which is also easily generalizable to create custom and hybrid layout algorithms.  
Our evaluation results demonstrate that the MARL graph drawing algorithms generate comparable results to their classical counterparts; with some amount of computational overhead.  
We hope this work will inspire more research in adapting RL for graph drawing. 

We end our paper with some further discussion on scalability. It is well known that most MARL algorithms have exponential computation complexity in the joint state-action space~\cite{ZhouYangChen2017}. 
That is, the size of the state-action space is exponentially large in the number of agents~\cite{QuWiermanLi2020}.
Since we utilize the MARL framework in graph drawing, our framework inherits the exponential complexity of MARL~\cite{BusoniuBabuskaSchutter2010} in terms of the number of nodes in a graph, making the graph drawing difficult to scale to really large graphs. 

However, not all hope is lost for scalability. 
Recent work by Qu \etal~\cite{QuWiermanLi2020, QuLinWierman2020} identified a class of networked MARL problems where ``the model exhibits a local dependence structure that allows it to be solved in a scalable manner". 
In particular, if a MARL model enforces local interactions, that is, agents are ``associated with a graph and they interact only with nearby agents in the graph"~\cite{QuLinWierman2020}, then it is possible to utilize the graph structure to develop scalable MARL algorithms~\cite{QuWiermanLi2020, QuLinWierman2020}.  
This \emph{localized policy} fits well with our MARL graph drawing framework. 
In this paper, we focus on showing the feasibility of MARL for graph drawing; we leave it for future work to address its scalability challenge.





\clearpage

\appendix
\section{Implementation Details}
\label{sec:implementation-details}

We provide additional implementational details for reproducibility. 
The implementation of the FR layout and stress majorization is provided by \textsf{Graphviz}~\cite{Gansner2014}.  
The DCG layout is available as part of \textsf{Cytoscape.js}. 
We implement MARL counterparts of these classical layout algorithms based on the formulation discussed in~\autoref{sec:interpretation} and by modifying and integrating the library \textsf{REINFORCEjs} with \textsf{Cytoscappe.js}. 
In particular, we implement both local and global stress for stress majorization. 
We make a distinction between parameter settings for each layout algorithms in general and the convergence criteria for comparative analysis in~\autoref{sec:evaluation}. 

\para{FR, DCG, Stress Majorization and their MARL counterparts.}
For FR and MARL FR algorithms, the optimal edge length $k=30$.
The parameters of {DGC} and {MARL DGC} algorithms include: $\lambda = 30$, $\zeta = 5$, and $\mu = 5000$. 
For {MARL Local Stress} algorithm, $p = 10$.

\para{MARL custom layouts.}
To derive new layout algorithms, we use formulation based on the local quality measure in~\autoref{eq:rewardFunction3} and implement the customized reward based on~\autoref{eq:quality}.
The default weight for each metric is determined using a grid search that maximizes the local quality measure for a fixed small graph: $\omega_1 = 0.35$, $\omega_2 = 0.20$, $\omega_3 = 0.10$, $\omega_4 = 0.25$, and $\omega_5 = 0.10$. 
The values of desired edge length and expected minimum node distance in {MARL Custom} layout algorithm are set to 30.

\para{MARL hybrid and incremental layouts.}
For our MARL hybrid implementation, the hybrid reward function is based on~\autoref{eq:hybrid} with $\beta = 0.5$.   
We also include a MARL-based incremental layout algorithm in {\tool}.  

\para{Convergence parameters.}
In our graph layout framework, there are four parameters to configure for the convergence criterion; the number of maximum iterations $M$, average node displacement $A$, displacement rate $\Delta A$, and the stress ratio $\delta E$.

Historically, Eades~\cite{Eades1984} asserted that all graphs reach to a minimal energy level after $100$ iterations. 
Dogrusoz \etal~\cite{DogrusozGiralCetintas2009} used $2500$ as the number of maximum iterations in their implementation.
We found that using the same value $M = 2500$ in our algorithms is a good balance between the maximum amount of running time and layout quality.
We determined the threshold values of average node displacement and displacement rate empirically as $A = 5$ and $\Delta A = 2$ (pixels).
Kamada and Kawai terminated their energy-based graph drawing when an energy threshold (referred to as a convergence precision) is reached~\cite{KamadaKawai1989}.
However, they did not provide information on how the threshold is determined.
Gansner \etal~\cite{GansnerKorenNorth2005} suggested $0.0001$ as a typical value for stress ratio tolerance.
We use the same suggested value for our algorithms, that is, $\delta E = 0.0001$.

All these parameter values achieve aesthetically pleasing layouts in a reasonable amount of time.
In addition, all convergence parameters can be configured both in the layout library API and from the user interface of the tool {\tool} itself.

\para{Node displacement with cooling schedule.}
Fruchterman and Reingold added the notion of ``temperature'' and ``cooling'' in their proposed algorithm~\cite{FruchtermanReingold1991}.
The maximum amount of node displacement is limited by the temperature, and the temperature is decreased over time by using a cooling factor.
The idea is that as the layout becomes better over time, smaller and smaller adjustments are needed, which is similar to a cooling process~\cite{FruchtermanReingold1991,Kobourov2012}.
Similar cooling schedule is used by  Davidson and Harel~\cite{DavidsonHarel1996},
and Dogrusoz \etal~\cite{DogrusozGiralCetintas2009}.

Davidson and Harel used a geometric rule for adjusting the temperature:
$
T_{t+1} = \tau * T_t, 
$
where $\tau$ is the cooling factor, and they set it to $0.75$ to achieve a relatively rapid cooling~\cite{DavidsonHarel1996}.
Their temperature reduction schedule is based on the input size.

In our work, we utilize the temperature and cooling for all MARL layout algorithms to determine how much a node moves when an action is taken.
Initially, each node can move by 10 pixels in each direction, and when the layout progresses the movement is gradually decreased.
The cooling rule is the same one used by Davidson and Harel~\cite{DavidsonHarel1996}. 
We reduce the temperature after every $|V| + |E|$ iterations.

\section{Qualitative Evaluation: Compute Results}
\label{sec:evaluation-details}

\begin{figure*}
    \centering
    \includegraphics[width=0.99\textwidth]{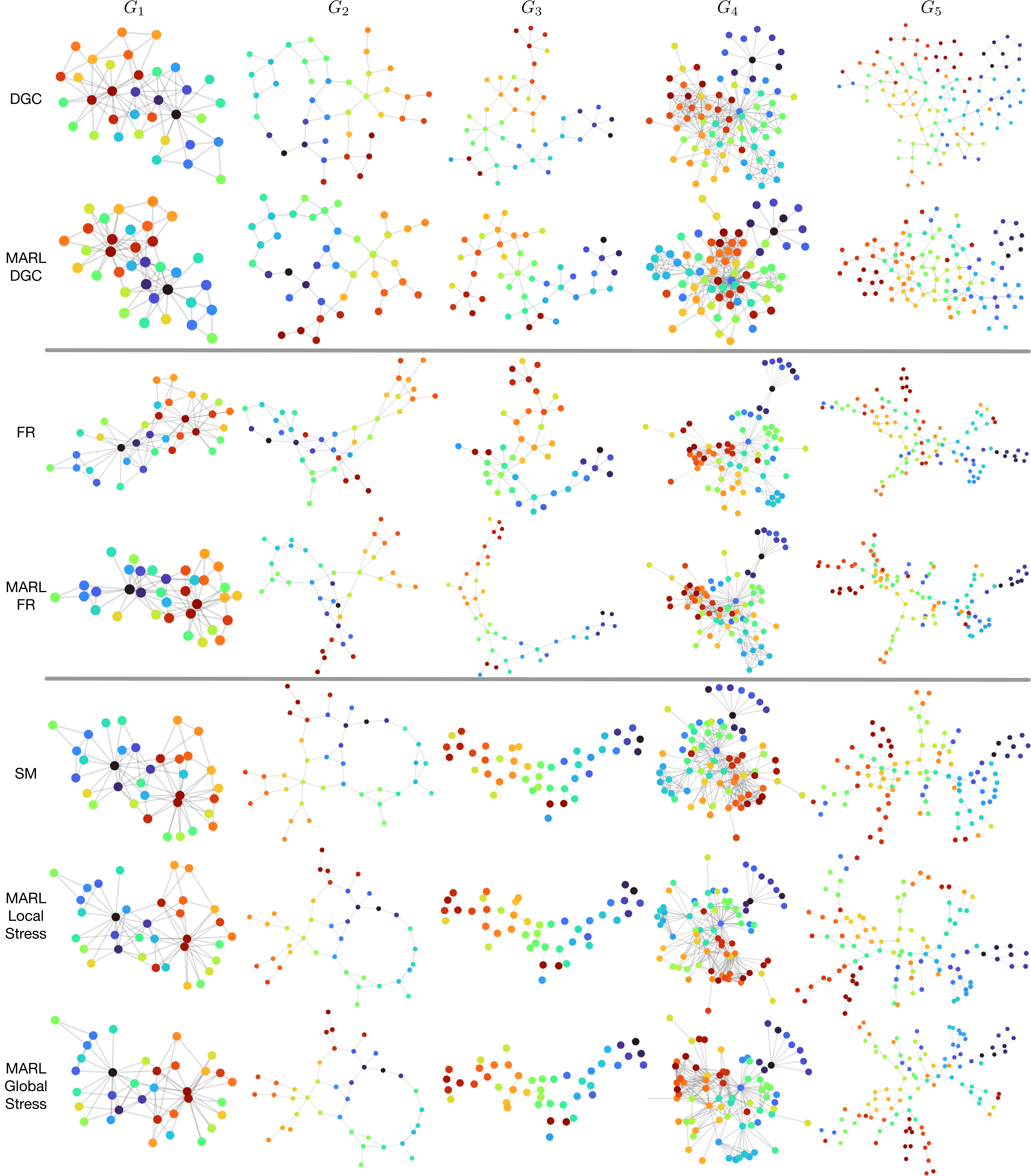}
    \caption{Qualitative results for $Q_1$, $Q_2$, $Q_3$, $Q_4$, and $Q_5$ for DGC layout, FR layout, stress majorization (SM), and their MARL versions.}
    \label{fig:evaluation-G1-G5-DGC-FR-SM}
\end{figure*}

\begin{figure*}
    \centering
    \includegraphics[width=0.99\textwidth]{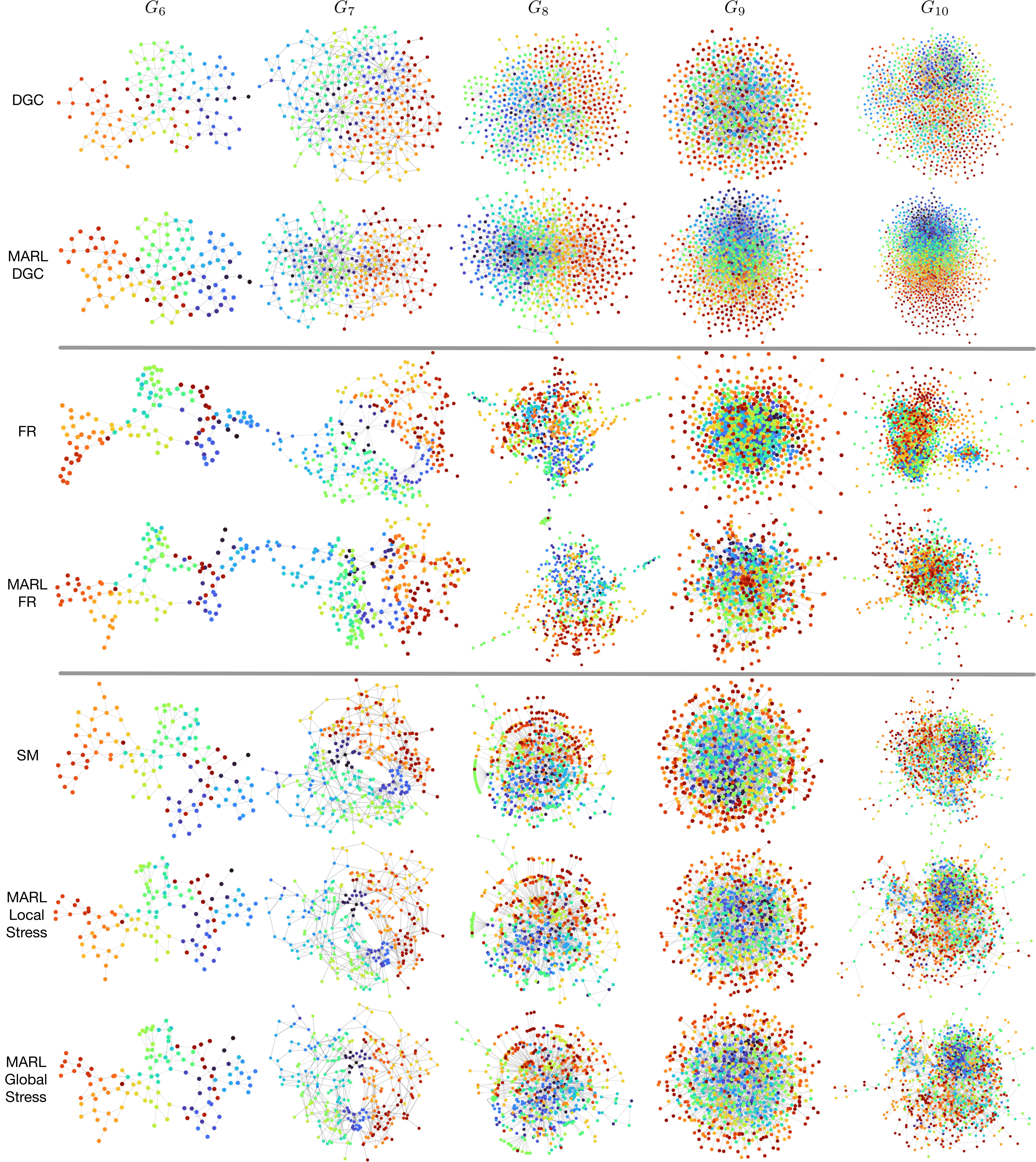}
    \caption{Qualitative results for $Q_6$, $Q_7$, $Q_8$, $Q_9$, and $Q_{10}$ for DGC layout, FR layout, stress majorization (SM), and their MARL versions.}
    \label{fig:evaluation-G6-G10-DGC-FR-SM}
\end{figure*}

\begin{figure*}
    \centering
    \includegraphics[width=0.99\textwidth]{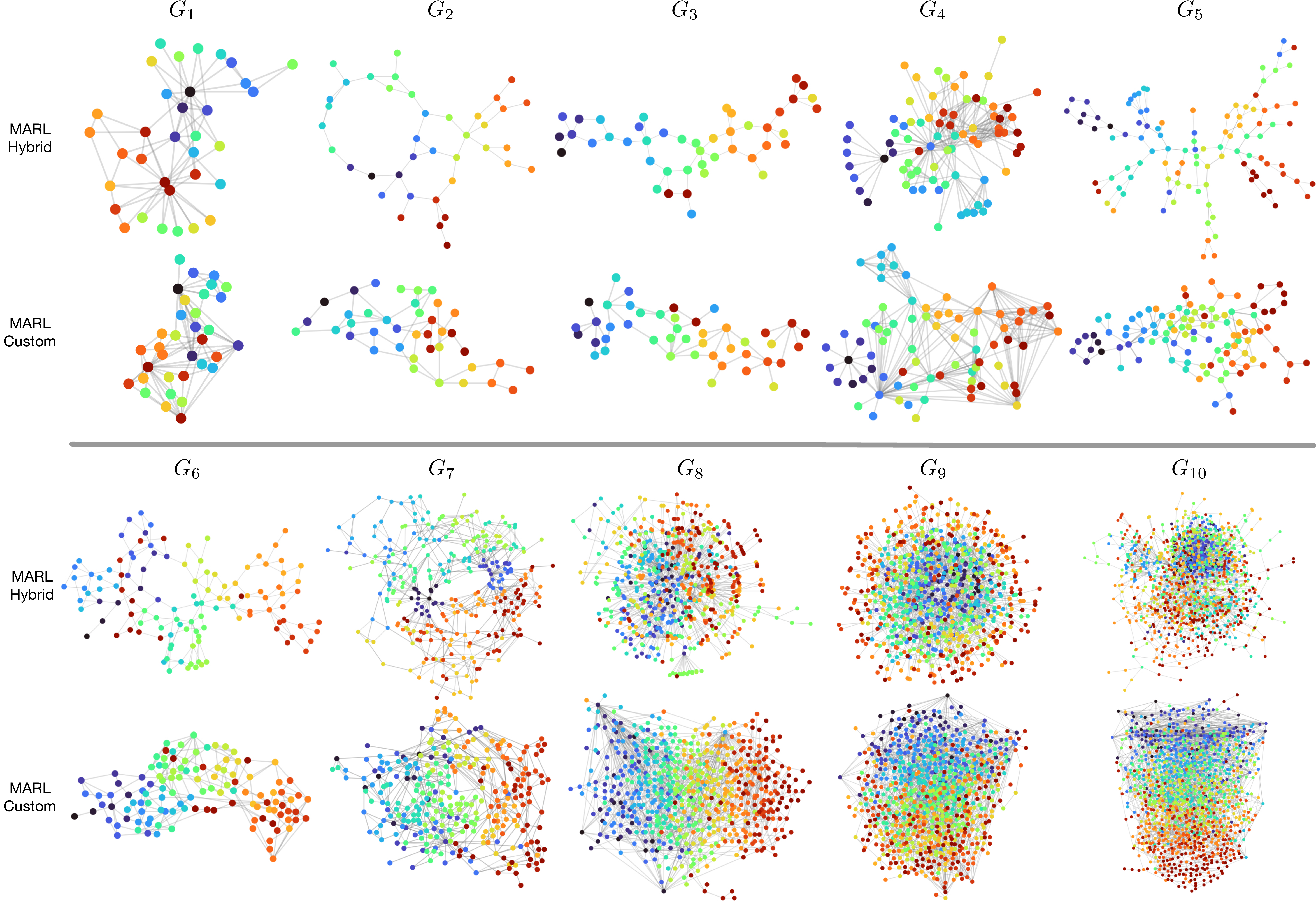}
    \caption{Qualitative results for $Q_1$ to $Q_{10}$ for MARL hybrid and MARL custom versions.}
    \label{fig:evaluation-All-hybrid-custom}
\end{figure*}

We provide complete results of the qualitative evaluation to demonstrate that the MARL framework produces aesthetically comparable results to the classical layout algorithms. 
We show examples $G_1$ to $G_5$ in~\autoref{fig:evaluation-G1-G5-DGC-FR-SM}, as well as $G_6$ to $G_{10}$ in~\autoref{fig:evaluation-G6-G10-DGC-FR-SM} where we compare the layouts obtained by classical algorithms (DGC, FR, and stress majorization) with those obtained by their MARL versions.  
We also include layouts obtained via a MARL hybrid and MARL custom for all graphs in~\autoref{fig:evaluation-All-hybrid-custom} . 
For each graph, nodes with the same color correspond with one another.

\end{document}